\documentclass[10pt,journal,compsoc]{IEEEtran}

\usepackage[nocompress]{cite}

\usepackage{url}
\usepackage{ragged2e}
\usepackage{epsfig}
\usepackage{graphicx}
\usepackage{amsmath,amssymb} 
\usepackage{mathptmx}      
\usepackage{subfigure}
\usepackage{algorithm}
\usepackage{algorithmicx}
\usepackage{algpseudocode}
\usepackage{graphics}
\usepackage{mathrsfs}
\usepackage{threeparttable}
\usepackage{color}
\usepackage[normalem]{ulem}
\usepackage{multirow}
\usepackage{float}
\usepackage{amsfonts}
\usepackage{bm}
\usepackage{array}
\usepackage[table]{xcolor}
\usepackage{colortbl}
\usepackage{pifont}
\usepackage{diagbox}
\usepackage{rotating}
\usepackage{booktabs}
\usepackage{overpic}
\usepackage{textcomp}
\usepackage{contour}
\usepackage{enumitem}
\usepackage{stfloats}
\usepackage{threeparttable}
\usepackage{makecell}
\usepackage[colorlinks=true,breaklinks=true,urlcolor=magenta]{hyperref}

\def\ie{\emph{i.e.}}
\def\eg{\emph{e.g.}}

\newcommand{\figref}[1]{Fig.~\ref{#1}}
\newcommand{\tabref}[1]{Table~\ref{#1}}

\newcommand{\secref}[1]{\S\ref{#1}}

\newcommand{\tgray}[1]{\textcolor{gray}{#1}}

\definecolor{mygray1}{gray}{.75}

\def\ie{\textit{i.e.}}
\def\eg{\textit{e.g.}}

\graphicspath{{./Imgs/}}
\DeclareGraphicsExtensions{.jpg,.pdf,.png}

\RequirePackage{silence}
\begin{document}

\title{SAM Struggles in Concealed Scenes – \\
Empirical Study on ``Segment Anything''}

\author{
Ge-Peng Ji,~
Deng-Ping Fan,~
Peng Xu,~
Ming-Ming Cheng,~
Bowen Zhou,~
Luc Van Gool\\
\IEEEcompsocitemizethanks{
\IEEEcompsocthanksitem Ge-Peng Ji is with the College of Engineering, Computing \& Cybernetics, ANU, Canberra, Australia.
\IEEEcompsocthanksitem Deng-Ping Fan and Luc Van Gool are with the Computer Vision Lab (CVL), ETH Zurich, Zurich, Switzerland.
\IEEEcompsocthanksitem Peng Xu and Bowen Zhou are with the Department of Electronic Engineering, Tsinghua University, Beijing, China.
\IEEEcompsocthanksitem Ming-Ming Cheng is with the Nankai University, Tianjin, China.
}
}


\IEEEtitleabstractindextext{%
\begin{abstract} \justifying
Segmenting anything is a ground-breaking step toward artificial general intelligence, and the Segment Anything Model (SAM) greatly fosters the foundation models for computer vision. We could not be more excited to probe the performance traits of SAM. In particular, exploring situations in which SAM does not perform well  is interesting. In this report, we choose three concealed scenes, \ie, camouflaged animals, industrial defects, and medical lesions, to evaluate SAM under unprompted settings. Our main observation is that SAM looks unskilled in concealed scenes.   
\end{abstract}

}

\maketitle

\IEEEdisplaynontitleabstractindextext

\IEEEpeerreviewmaketitle

\IEEEraisesectionheading{\section{Introduction}\label{sec:introduction}}

\IEEEPARstart{L}{arge} models open up new opportunities for artificial intelligence. In the past few months, there has been a boom in training foundation models on the vast linguistic corpus to produce amazing applications,
\eg, ChatGPT\footnote{\url{https://chat.openai.com}}, GPT-4\footnote{\url{https://openai.com/research/gpt-4}}.
Both natural language processing and multimodal learning communities have been revolutionized.
Large models' capacity for generalization and emergent makes it easy for users to believe that large models can solve anything.

Last week, 
the ``Segment Anything''~\cite{kirillov2023segment} project was released, and its Segment Anything Model (SAM) is a large ViT-based model trained on the large visual corpus (SA-1B).
This is a ground-breaking step toward artificial general intelligence, as SAM demonstrates promising segmentation capabilities in various scenarios and the great potential of the  foundation models for computer vision. Like all computer vision researchers, we cannot wait to probe the performance traits of SAM to help the community to comprehend it further. Moreover, it is interesting to explore the situations in which SAM does not work well.

In this paper, we compare SAM quantitatively with cutting-edge models on camouflaged object segmentation task, and present diversified visualization results in three concealed scenes\footnote{There is a newly released survey on deep concealed scene understanding~\cite{fan2023advances} (Project link: \url{https://github.com/DengPingFan/CSU}).}~\cite{fan2023advances}, \ie, camouflaged animals, industrial defects, and medical lesions. Our main observation is that SAM looks not skillful in concealed scenes.

\section{Experiment}\label{sec:experiment}

We use three frequently-used  camouflaged object segmentation (COS) benchmarks to evaluate SAM. Next, we describe our mask selection strategy (\secref{sec:mask_selection_strategy}) and evaluation protocols (\secref{sec:evaluation_protocols}), compare SAM with the state-of-the-art models of COS (\secref{sec:quantitative_eval}), and provide qualitative visualisations on  three concealed scenarios (\secref{sec:qualitative_eval}).

\subsection{Mask Selection Strategy}\label{sec:mask_selection_strategy}

If under the unprompted setting, SAM generates multiple binary masks and can pop out several potential objects within an input. 
For a fair evaluation of interesting regions in a specific segmentation task, we take a strategy to select the most appropriate mask based on its ground-truth mask. Formally, given $N$ binary predictions $\{\mathbf{P}_n\}_{n=1}^{N}$ and the ground-truth $\mathbf{G}$ for an input image, we calculate intersection over union (IoU) scores for each pair to generate a set of evaluation scores $\{ \text{IoU}_n \}_{n=1}^N$. We finally select the mask with the highest IoU score from this set.

\subsection{Evaluation Protocols}\label{sec:evaluation_protocols}

Our protocols are  following the standard practice as in~\cite{fan2022concealed}.

\noindent$\bullet$~\textbf{Datasets.} We use three commonly-used COS benchmarks in our experiments, including CAMO~\cite{le2019anabranch} (250 samples), COD10K~\cite{fan2020camouflaged} (2,026 samples), and NC4K~\cite{lv2021simultaneously} (4,121 samples).

\noindent$\bullet$~\textbf{Models.} To ensure a fair comparison with SAM, we choose the current top-performing COS models using transformer architecture, \ie, CamoFormer-P/-S~\cite{yin2023camoformer}, HitNet~\cite{hu2023high}.

\noindent$\bullet$~\textbf{Metrics.} We use five commonly-used metrics for the evaluation: structure measure ($S_{\alpha}$)~\cite{fan2017structure}, enhanced-alignment measure ($E_{\phi}$)~\cite{fan2021cognitive}, F-measure ($F_{\beta}$)~\cite{borji2015salient,zhuge2022salient}, weighted F-measure ($F_{\beta}^{w}$)~\cite{margolin2014evaluate}, and mean absolute error ($M$). According to different thresholding strategies, the adaptive/mean/maximum values of F-measure and E-measure are reported.
We denote different $E_\phi$ scores as $E_\phi^{ad}$, $E_\phi^{mn}$, and $E_\phi^{mx}$.

\subsection{Quantitative Evaluation}\label{sec:quantitative_eval}

We report the quantitative comparison in~\tabref{tab:cos_benchmark_cod10k}, SAM demonstrates significant improvements as model capabilities increase from ViT-B to ViT-L, with an increase in $F_{\beta}^{w}$ score from $0.353$ to $0.655$ on CAMO. However, the improvement is limited when the model becomes larger, increasing only from $0.655$ (ViT-L) to $0.700$ (ViT-H). 
Moreover, we observe that there remains a large gap between SAM even with ViT-H and current top-performing COS models on three benchmarks. For example as presented in \tabref{tab:cos_benchmark_cod10k}, the difference of $E_\phi^{mx}$ score between SAM (ViT-H) and CamoFormer-S~\cite{yin2023camoformer} reaches 13.8\% on COD10K dataset, 25.6\% on CAMO dataset, and 16.9\% on NC4K dataset. This gap indicates that the perception ability of SAM needs further improvement for concealed scenes.

\begin{table*}[t!]
\centering
\caption{\textbf{Quantitative comparison on three popular COS benchmarks.} The symbols $\uparrow$/$\downarrow$ indicate that a higher/lower score is better. The highest scores are marked in bold. $\Delta$ represents the difference between SAM and the highest score achieved by current cutting-edge COS models.}
\vspace{-8pt}
\label{tab:cos_benchmark_cod10k}
\footnotesize
\renewcommand{\arraystretch}{1}
\renewcommand{\tabcolsep}{0.20cm}
\begin{threeparttable}
\begin{tabular}{| c | r | r | r || r|r|r|rrr|rrr| }
    \hline
    \rowcolor{mygray1}
    &Model~ &Pub/Year &Backbone~ &$S_{\alpha}\uparrow$ &$F_\beta^w\uparrow$ &$M\downarrow$ &$E_\phi^{ad}\uparrow$ &$E_\phi^{mn}\uparrow$ &$E_\phi^{mx}\uparrow$ &$F_\beta^{ad}\uparrow$ &$F_\beta^{mn}\uparrow$ &$F_\beta^{mx}\uparrow$ \\
    \hline
    \hline
    \multirow{9}{*}{\begin{sideways}COD10K~\cite{fan2020camouflaged}\end{sideways}} &CamoFormer-P~\cite{yin2023camoformer} &arXiv$_{23}$ &PVTv2-B4~\cite{wang2022pvt} 
    &0.869 &0.786 &\textbf{0.023} &0.931 &0.932 &0.939 &0.794 &0.811 &0.829 \\
    &CamoFormer-S~\cite{yin2023camoformer} &arXiv$_{23}$ &Swin-B~\cite{liu2021swin} 
    &0.862 &0.772 &0.024 &0.932 &0.931 &\textbf{0.941} &0.780 &0.799 &0.818 \\
    &HitNet~\cite{hu2023high} &AAAI$_{23}$ &PVTv2-B2~\cite{wang2022pvt} 
    &\textbf{0.871} &\textbf{0.806} &\textbf{0.023} &\textbf{0.936} &\textbf{0.935} &0.938 &\textbf{0.818} &\textbf{0.823} &\textbf{0.838} \\
    \cline{2-13}
    &\multirow{6}{*}{SAM~\cite{kirillov2023segment}} &\multirow{6}{*}{arXiv$_{23}$} &ViT-B~\cite{vit2021ICLR} &0.585 &0.353 &0.108 &0.535 &0.533 &0.535 &0.423 &0.422 &0.423 \\
    &&&\tgray{Difference ($\Delta$)} &\tgray{-28.6\%} &\tgray{-45.3\%} &\tgray{+8.5\%} &\tgray{-40.1\%} &\tgray{-40.2\%} &\tgray{-40.3\%} &\tgray{-39.5\%} &\tgray{-40.1\%} &\tgray{-41.5\%} \\
    &&&ViT-L~\cite{vit2021ICLR} &0.751 &0.655 &0.065 &0.766 &0.764 &0.766 &0.718 &0.716 &0.718\\
    &&&\tgray{Difference ($\Delta$)} &\tgray{-12\%} &\tgray{-15.1\%} &\tgray{+4.2\%} &\tgray{-17\%} &\tgray{-17.1\%} &\tgray{-17.2\%} &\tgray{-10\%} &\tgray{-10.7\%} &\tgray{-12\%} \\
    &&&ViT-H~\cite{vit2021ICLR} &0.781 &0.700 &0.054 &0.800 &0.798 &0.800 &0.756 &0.754 &0.756 \\
    &&&\tgray{Difference ($\Delta$)} &\tgray{-9\%} &\tgray{-10.6\%} &\tgray{+3.1\%} &\tgray{-13.6\%} &\tgray{-13.7\%} &\tgray{-13.8\%} &\tgray{-6.2\%} &\tgray{-6.9\%} &\tgray{-8.2\%} \\
    \hline
    \hline
    \rowcolor{mygray1}
    &Model~ &Pub/Year &Backbone~ &$S_{\alpha}\uparrow$ &$F_\beta^w\uparrow$ &$M\downarrow$ &$E_\phi^{ad}\uparrow$ &$E_\phi^{mn}\uparrow$ &$E_\phi^{mx}\uparrow$ &$F_\beta^{ad}\uparrow$ &$F_\beta^{mn}\uparrow$ &$F_\beta^{mx}\uparrow$ \\
    \hline
    \hline
    \multirow{9}{*}{\begin{sideways}CAMO~\cite{le2019anabranch}\end{sideways}}  &CamoFormer-P~\cite{yin2023camoformer} &arXiv$_{23}$ &PVTv2-B4~\cite{wang2022pvt} &0.872 &0.831 &0.046 &0.931 &0.929 &\textbf{0.938} &0.853 &0.854 &0.868 \\
    &CamoFormer-S~\cite{yin2023camoformer} &arXiv$_{23}$ &Swin-B~\cite{liu2021swin} &\textbf{0.876} &\textbf{0.832} &\textbf{0.043} &\textbf{0.935} &\textbf{0.930} &\textbf{0.938} &\textbf{0.856} &\textbf{0.856} &\textbf{0.871} \\
    &HitNet~\cite{hu2023high} &AAAI$_{23}$ &PVTv2-B2~\cite{wang2022pvt} &0.849 &0.809 &0.055 &0.910 &0.906 &0.910 &0.833 &0.831 &0.838 \\
    \cline{2-13}
    &\multirow{6}{*}{SAM~\cite{kirillov2023segment}} &\multirow{6}{*}{arXiv$_{23}$} &ViT-B~\cite{vit2021ICLR} &0.462 &0.238 &0.219 &0.402 &0.401 &0.402 &0.312 &0.312 &0.312 \\
    &&&\tgray{Difference ($\Delta$)}  &\tgray{-41.4\%} &\tgray{-59.4\%} &\tgray{+17.6\%} &\tgray{-53.3\%} &\tgray{-52.9\%} &\tgray{-53.6\%} &\tgray{-54.4\%} &\tgray{-54.4\%} &\tgray{-55.9\%} \\
    &&&ViT-L~\cite{vit2021ICLR} &0.630 &0.534 &0.162 &0.628 &0.626 &0.628 &0.617 &0.615 &0.617 \\
    &&&\tgray{Difference ($\Delta$)}  &\tgray{-24.6\%} &\tgray{-29.8\%} &\tgray{+11.9\%} &\tgray{-30.7\%} &\tgray{-30.4\%} &\tgray{-31\%} &\tgray{-23.9\%} &\tgray{-24.1\%} &\tgray{-25.4\%} \\
    &&&ViT-H~\cite{vit2021ICLR} &0.677 &0.594 &0.136 &0.682 &0.680 &0.682 &0.670 &0.668 &0.670 \\
    &&&\tgray{Difference ($\Delta$)}  &\tgray{-19.9\%} &\tgray{-23.8\%} &\tgray{+9.3\%} &\tgray{-25.3\%} &\tgray{-25\%} &\tgray{-25.6\%} &\tgray{-18.6\%} &\tgray{-18.8\%} &\tgray{-20.1\%} \\
    \hline
    \hline
    \rowcolor{mygray1}
    &Model~ &Pub/Year &Backbone~ &$S_{\alpha}\uparrow$ &$F_\beta^w\uparrow$ &$M\downarrow$ &$E_\phi^{ad}\uparrow$ &$E_\phi^{mn}\uparrow$ &$E_\phi^{mx}\uparrow$ &$F_\beta^{ad}\uparrow$ &$F_\beta^{mn}\uparrow$ &$F_\beta^{mx}\uparrow$ \\
    \hline
    \hline
    \multirow{9}{*}{\begin{sideways}NC4K~\cite{lv2021simultaneously}\end{sideways}} &CamoFormer-P~\cite{yin2023camoformer} &arXiv$_{23}$ &PVTv2-B4~\cite{wang2022pvt} &\textbf{0.892} &\textbf{0.847} &\textbf{0.030} &\textbf{0.941} &\textbf{0.939} &\textbf{0.946} &\textbf{0.863} &\textbf{0.868} &\textbf{0.880} \\
    &CamoFormer-S~\cite{yin2023camoformer} &arXiv$_{23}$ &Swin-B~\cite{liu2021swin} &0.888 &0.840 &0.031 &\textbf{0.941} &0.937 &\textbf{0.946} &0.857 &0.863 &0.877 \\
    &HitNet~\cite{hu2023high} &AAAI$_{23}$ &PVTv2-B2~\cite{wang2022pvt} &0.875 &0.834 &0.037 &0.928 &0.926 &0.929 &0.854 &0.853 &0.863 \\
    \cline{2-13}
    &\multirow{6}{*}{SAM~\cite{kirillov2023segment}} &\multirow{6}{*}{arXiv$_{23}$} &ViT-B~\cite{vit2021ICLR} &0.544 &0.334 &0.166 &0.494 &0.493 &0.494 &0.403 &0.403 &0.403 \\
    &&&\tgray{Difference ($\Delta$)} &\tgray{-34.8\%} &\tgray{-51.3\%} &\tgray{+13.6\%} &\tgray{-44.7\%} &\tgray{-44.6\%} &\tgray{-45.2\%} &\tgray{-46\%} &\tgray{-46.5\%} &\tgray{-47.7\%} \\
    &&&ViT-L~\cite{vit2021ICLR} &0.728 &0.643 &0.101 &0.735 &0.733 &0.735 &0.706 &0.704 &0.706 \\
    &&&\tgray{Difference ($\Delta$)}  &\tgray{-16.4\%} &\tgray{-20.4\%} &\tgray{+7.1\%} &\tgray{-20.6\%} &\tgray{-20.6\%} &\tgray{-21.1\%} &\tgray{-15.7\%} &\tgray{-16.4\%} &\tgray{-17.4\%} \\
    &&&ViT-H~\cite{vit2021ICLR} &0.763 &0.696 &0.087 &0.777 &0.775 &0.777 &0.752 &0.750 &0.752\\
    &&&\tgray{Difference ($\Delta$)} &\tgray{-12.9\%} &\tgray{-15.1\%} &\tgray{+5.7\%} &\tgray{-16.4\%} &\tgray{-16.4\%} &\tgray{-16.9\%} &\tgray{-11.1\%} &\tgray{-11.8\%} &\tgray{-12.8\%} \\
    \hline
\end{tabular}
\end{threeparttable}
\end{table*}

\begin{figure*}[ht!]
  \centering
  \vspace{-5pt}
  \includegraphics[width=\linewidth]{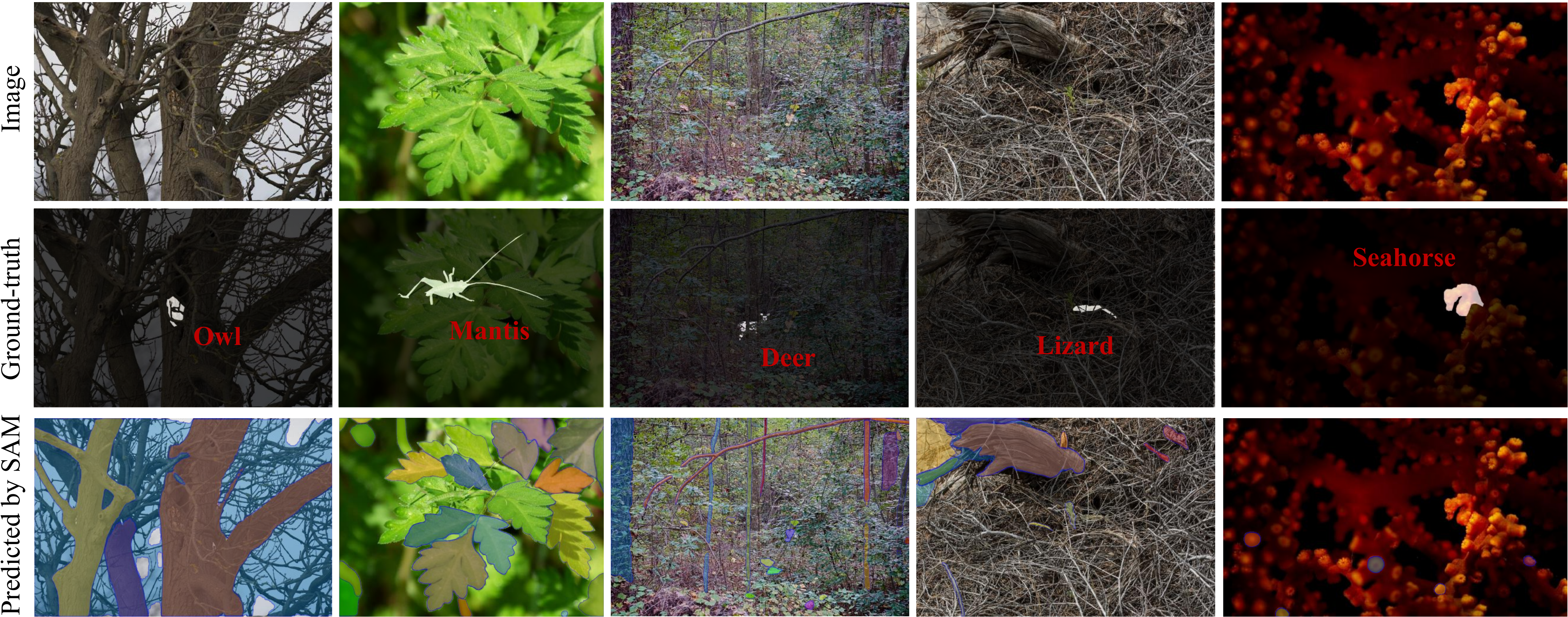}
  \vspace{-20pt}
  \caption{SAM~\cite{kirillov2023segment} fails to perceive the animals that are visually ``hidden'' in their natural surroundings. 
 All the samples are from COD10K dataset~\cite{fan2020camouflaged}.
 }
 \vspace{-10pt}
  \label{fig:teaser_cos}
\end{figure*}

\begin{figure*}[pt!]
  \centering
  \includegraphics[width=\linewidth]{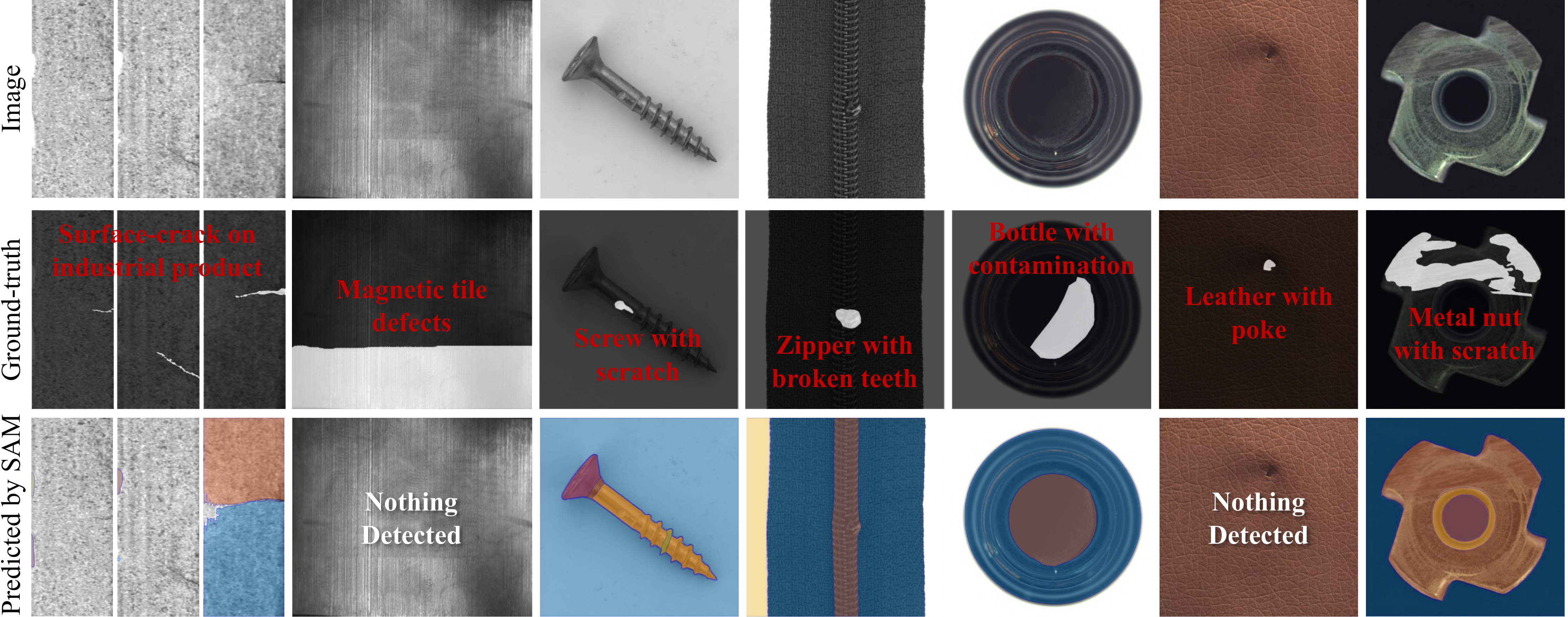}
  \vspace{-20pt}
  \caption{
   SAM~\cite{kirillov2023segment} is unskilled in detecting concealed defects in industrial scenes. 
   These samples are taken from KolektorSDD~\cite{tabernik2020segmentation}, MagneticTile~\cite{huang2020surface}, and MVTecAD~\cite{bergmann2021mvtec} datasets.}
   \vspace{-8pt}
  \label{fig:teaser_ids}
\end{figure*}

\begin{figure*}[pt!]
  \centering
  \includegraphics[width=\linewidth]{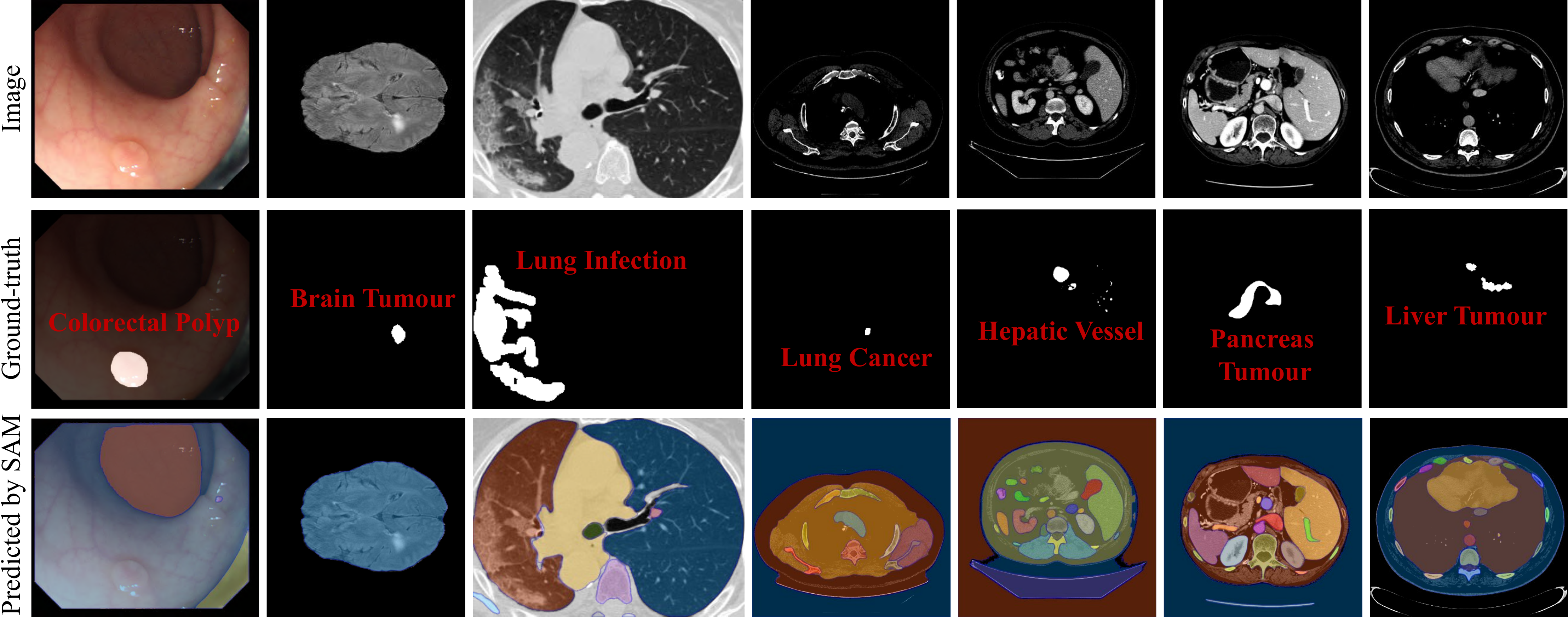}
  \vspace{-20pt}
  \caption{
 SAM~\cite{kirillov2023segment} fails to detect these lesion regions in various medical modalities. These samples cover the RGB colour modality from CVC-300~\cite{bernal2012towards} (1\textsuperscript{st} column); the MRI modality from BraTS2021~\cite{baid2021rsna} (2\textsuperscript{nd} column); the CT modalities from COVID-SemiSeg~\cite{fan2020infnet} (3\textsuperscript{rd} column) and MSD~\cite{antonelli2022medical} (from 4\textsuperscript{th} to 7\textsuperscript{th} columns).
  }
  \vspace{-10pt}
  \label{fig:teaser_mis}
\end{figure*}

\subsection{Qualitative Comparison}\label{sec:qualitative_eval}

We further qualitatively evaluate SAM in three concealed scenarios, and several interesting findings are as follows. {All the visualisation results are generated by the \href{https://segment-anything.com/demo}{online demo} of SAM.}
\textit{a) Camouflaged animal.} As presented in~\figref{fig:teaser_cos}, it is difficult for SAM to detect concealed animals in their natural habitat. SAM fails to segment a mantis crouching on a leaf (in the second column) and a seahorse swimming in an orange coral (in the last column). In these two cases, SAM struggles to distinguish the target semantics from their surroundings because the foreground and background share similar appearances of shape and colour. As a result, SAM becomes more dependent on unreliable pixel intensity changes along the boundaries.
\textit{b) Industrial defect.} In this scenario, given a  product, it is usually collected by the  close-range shots, occupying a large area in the image, and SAM's behaviour appears to segment the main part of the object, such as the screw in the 3\textsuperscript{rd} column and bottle in the 5\textsuperscript{th} column of \figref{fig:teaser_ids}. Furthermore, we notice that it is difficult for SAM to distinguish defective areas from the textured background. For instance, products with surface cracks (in 1\textsuperscript{st} column) and leather with a poke (in the 6\textsuperscript{th} column) are challenging to identify accurately. This phenomenon is not surprising because SAM is trained on natural objects with standard sizes and high-contrast attributes.
\textit{c) Medical lesion.} As illustrated in 1\textsuperscript{st} column of~\figref{fig:teaser_mis}, we observe that SAM does not handle medical data with concealed patterns well, such as benign colorectal polys that share similar colours with the surrounding tissues. The remaining samples in~\figref{fig:teaser_mis} are grayscale slices from {three-dimensional MRI and CT scans}. SAM can roughly segment the organ regions since they have distinct boundaries, but it does not perform well in recognizing amorphous lesion regions, \eg, cancer, vessels, and tumours. This suggests that SAM lacks the medical domain knowledge of these anatomical and pathological cases. To alleviate this limitation,  the intrinsic relationships and semantics of anatomical structures can be injected into SAM, such as the assumption that liver tumours should be inside the liver, rather than the brain.

\subsection{Discussion}

From the above empirical analyses, our conclusions are:

\noindent~$\bullet$ We observe that SAM often segments an occluded object into multiple separated masks, indicating that its semantic capabilities in concealed scenes can be improved.

\noindent~$\bullet$ Unlike self-supervised large language models, SAM employs supervised training, in our experiments its emergent and reasoning abilities have not been observed. Thus, it would be interesting to try if more challenging training tasks improve its performance.

\noindent~$\bullet$ Considering the practical open-set problem, now the granularity and uncertainty are the bottlenecks of SAM, limiting its applications to the scenes that require high accuracy, \eg, autonomous driving and clinical diagnosis. To alleviate this issue, one potential solution is to support the model with prior knowledge.

\noindent~$\bullet$ SAM's great success is demonstrating the power of data-centric AI in the large model era. We see a significant trend that human feedback-based learning and large foundation models bring new opportunities for the vision community.

\section{Conclusion}\label{sec:conclusion}

In summary, this paper presents an empirical study of SAM. Firstly, we quantitatively evaluate SAM using cutting-edge models on the camouflaged object segmentation task. Secondly, we present several failure cases in three concealed scenarios: camouflaged animals, industrial defects, and medical lesions. We expect that this paper helps the readers to comprehend SAM's performance traits in concealed scenes and brings new ideas to computer vision researchers.

\ifCLASSOPTIONcaptionsoff
  \newpage
\fi

{
\bibliographystyle{IEEEtran}
\bibliography{bibliography}
}

\vfill

\end{document}